\documentclass[journal]{IEEEtran}

\ifCLASSINFOpdf
\else
\fi

\usepackage{epsfig} 
\usepackage{cite}
\usepackage{graphicx}
\usepackage[cmex10]{amsmath}
\usepackage{algorithmic}
\usepackage{algorithm}
\usepackage{array}
\usepackage{mdwmath}
\usepackage{mdwtab}
\usepackage{eqparbox}
\usepackage[export]{adjustbox}
\usepackage{subfig}
\usepackage{url}
\usepackage{amsmath,amsfonts}
\usepackage{ctable}
\usepackage{color,soul}

\begin{document}

\title{A Large Force Haptic Interface\\with Modular Linear Actuators}
\author{Yeongtae~Jung$^1$* and~Joao~Ramos$^2$
\thanks{$^1$The author is with the Department of Mechanical System Engineering at the Jeonbuk National University, Jeonju, 54896, Republic of Korea.}
\thanks{$^2$The author is with the Department of Mechanical Science and Engineering at the University of Illinois at Urbana-Champaign, Urbana, IL 61801, USA.}
\thanks{*Corresponding author. Email: ytjung@jbnu.ac.kr}
\thanks{This work is supported by a gift from Google and by the National Science Foundation via grant IIS-2024775.}
}

\maketitle

\begin{abstract}
This paper presents a haptic interface with modular linear actuators which can address limitations of conventional devices based on rotatory joints. The proposed haptic interface is composed of parallel linear actuators that provide high backdrivability and small inertia. The performance of the haptic interface is compared with the conventional mechanisms in terms of force capability, reflected inertia, and structural stiffness. High stiffness, large range of motion with high force capability are achieved with the proposed mechanism, which are in trade-off relationships in traditional haptic interfaces. The device can apply up to 83 N continuously, which is three times larger than most haptic devices. The theoretical minimum haptic force density and the stiffness of the proposed mechanism were 1.3 to 1.9 times and 37 times of conventional mechanisms in a similar condition, respectively. The system is also scalable because its structural stiffness only depends on the timing belt stiffness, while that of conventional haptic interfaces is inversely proportional to the cube of structural lengths. The modular actuator design enables change of degrees freedom (DOFs) for different applications. The proposed haptic interface was tested by the interaction experiment with a virtual environment with rigid walls.

\end{abstract}

\IEEEpeerreviewmaketitle

\section{Introduction}
A haptic interface provides physical interaction between a user and a virtual environment. It's hardware and controller decide how well the system can render an environment to the user transparently in a stable manner. Although advanced control algorithms can enhance the performance of the haptic interface~\cite{JRNL:Ryu_Haptic_Passivity,JRNL:Khatib_Haptic,JRNL:Sirouspour_Haptics_Transparency,JRNL:Pan_Haptics_Transparency}, transparency and stability are conflicting objectives in the sense of control~\cite{JRNL:Lawrence_Tele_Passivity}. Thus, the fundamental limitation of a haptic system is determined by the physical system of the device, such as actuator bandwidth, structural stiffness, friction, inertia, and range of motion (ROM). The design of haptic interface hardware always introduces trade-offs between them. Large ROM requires long linkage lengths, which result in low structural stiffness or large reflected inertia from the linkages. Large force capability requires high gearing ratios which shows large friction and reflected inertia \cite{SimICRA21}. These trade-off relationships are dominated by the actuators, transmissions, and sensors of the haptic interface.

Serial mechanisms with rotational actuators, which is a widely used combination for robotic arms and exoskeleton systems, can be considered as the mechanism of a haptic interface~\cite{JRNL:L-EXOS,CONF:DLRHaptics,JRNL:HIROIII}. This mechanism is simple and can enable large range of motion since the interference between linkages can be prevented by design. However, it has disadvantages in displayable stiffness since the deformation of the linkages, which is mostly caused by the bending of the linkages, cumulatively affects the end-effector stiffness. Thus, it is suffered from large inertia of the linkages that is required for high stiffness of the system. This also introduces gearing of the actuator to handle the large inertia which negatively affects to the backdrivability of the haptic interface. 

\begin{figure}
\centering 
    \includegraphics[width = \columnwidth]{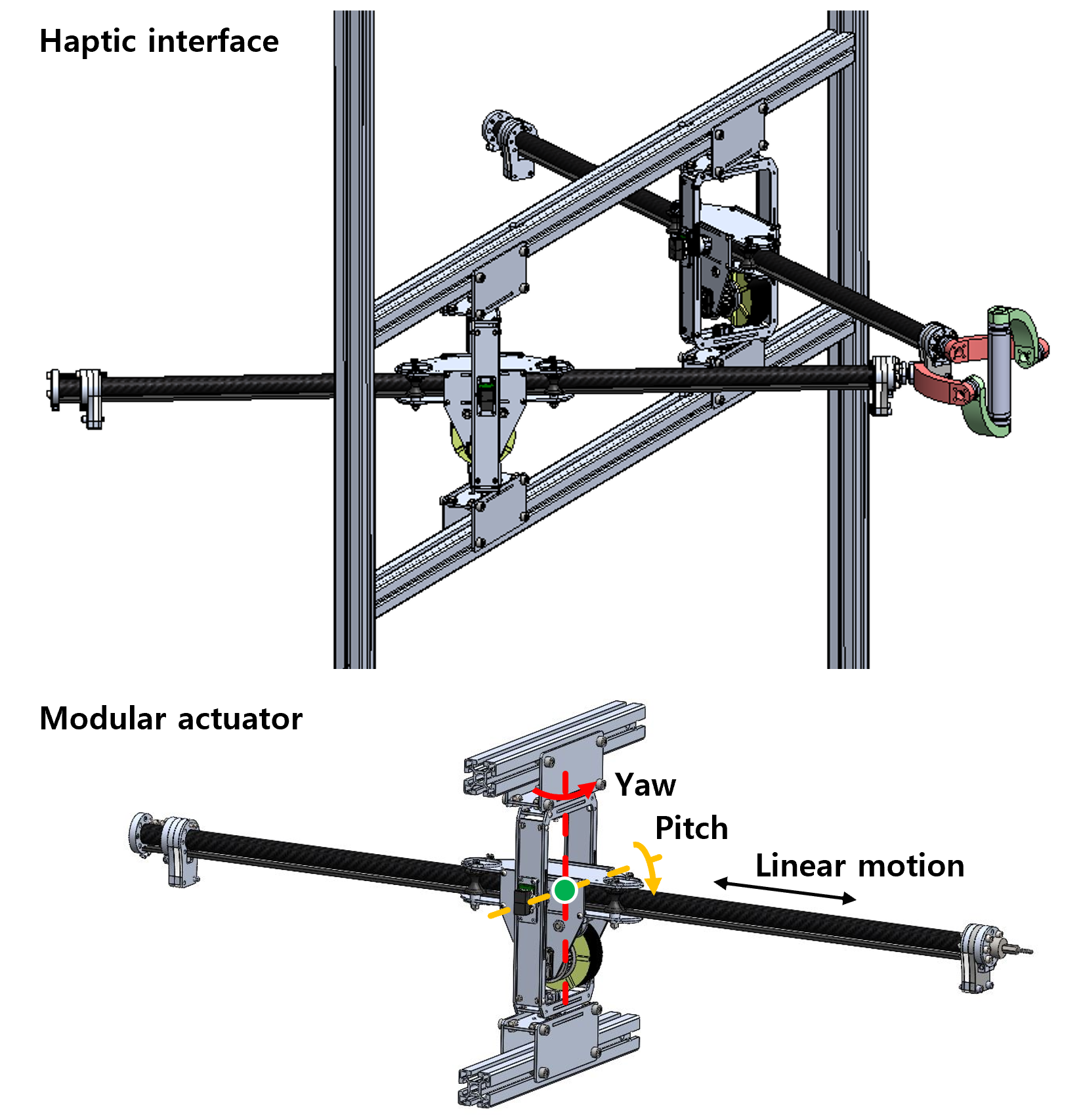}
    \caption{Proposed planar linear haptic interface with modular actuators.}
    \label{fig:Haptic_Interface}
\end{figure}

For this reason, most commercial haptic interfaces utilize parallel linkag-based mechanisms such as delta~\cite{WEB:Delta} or four-bar linkages~\cite{WEB:Phantom,WEB:HD2,WEB:Cyberforce}, with geared motors. This approach can enhance the structural stiffness of the system with parallel linkages, but limits the range of motion because of its structural complexity. Also, all the commercial haptic interfaces cited above shows force capability less than 20 N. This is because high reduction ratio gears were avoided to achieve high transparency of the system. Admittance control with force feedback could be an option for friction from the gearing. This design enables high force and lightweight designs with small actuators. However, the haptic feedback bandwidth and stability are limited by the sensing rate of the system and the noise level of the sensor, which requires high cost. The high gear ratio also reduces the available speed of the end-effector, which affects the force bandwidth. Researchers in \cite{WEB:Inca} utilized a tendon-driven haptic interface to enable a room-size range of motion. However, the stiffness is limited by the long tendon, and the tendons can interfere with the user.

Some non-commercialized haptic interfaces were proposed aiming to achieve high force capability, a large range of motion, and high back-drivability~\cite{CONF:DM2,JRNL:VirtuaPower,CONF:Mantis}. A hybrid actuation mechanism was introduced separating high- and low-frequency force generation with a large series elastic actuation at the base part, and a small, fast actuator at the distal joint side in~\cite{CONF:DM2}. However, this approach introduces massive complexity on the system together with the control efforts in contact with high stiffness virtual objects. Lee $\textit{et al}$. utilized parallel four-bar linkages that can provide a large range of motion and high force capability and stiffness~\cite{JRNL:VirtuaPower}. However, high reduction ratio gears were used to enlarge the force capability which can cause large reflected inertia of the actuator at the end-effector side, together with the friction. As the result, it showed large overshoot and long settling time. Barnaby $\textit{et al}$. developed a haptic interface with back-drivable actuators and timing belt reductions, together with a four-bar linkage mechanism~\cite{CONF:Mantis}. The system showed high backdrivabilty, but it could provide only a small force as well as commercial haptic interfaces.

The point to note is that all of the above-mentioned haptic interfaces use rotational actuation mechanisms. This introduces \textit{bending} deflection of the linkages, which is a dominant deflection of the structure. Even though the actuators themselves can generate high stiffness, low structural stiffness limits the displayable stiffness at the end-effector.

In this paper, we propose a novel haptic interface based on modular linear actuators that can provide large force, large ROM with mechanical backdrivability with small reflected inertia and high structural stiffness (Fig.~\ref{fig:Haptic_Interface}). The mechanism of the proposed haptic interface design is compared with conventional serial and parallel mechanisms in terms of available force over reflected inertia. The proposed mechanism is scalable since high structural stiffness is ensured by the linear actuator-based mechanism regardless of the ROM, which also results in small reflected inertia. The actuator provides high force capability with backdrivability because of the timing belt-based reduction. A gimbal handle design that can connect two modular linear actuators and ensure zero residual torque is introduced as an example of the proposed haptic interface design. The rest of this article is organized as follows. In Section II, the reflected inertia and stiffness of the proposed linear actuator-based mechanism are analyzed together with conventional rotational mechanisms. A 2-DOF prototype haptic interface with high-force and backdrivable linear actuators is introduced in Section III. The performance of the linear actuator and experimental result with a virtual object is presented in section IV. Section V concludes this article with discussions.

\section{Mechanism design}

\begin{figure}
    \centering 
    \subfloat[Overview of the system.]{\label{fig:Overview}\includegraphics[width = \columnwidth]{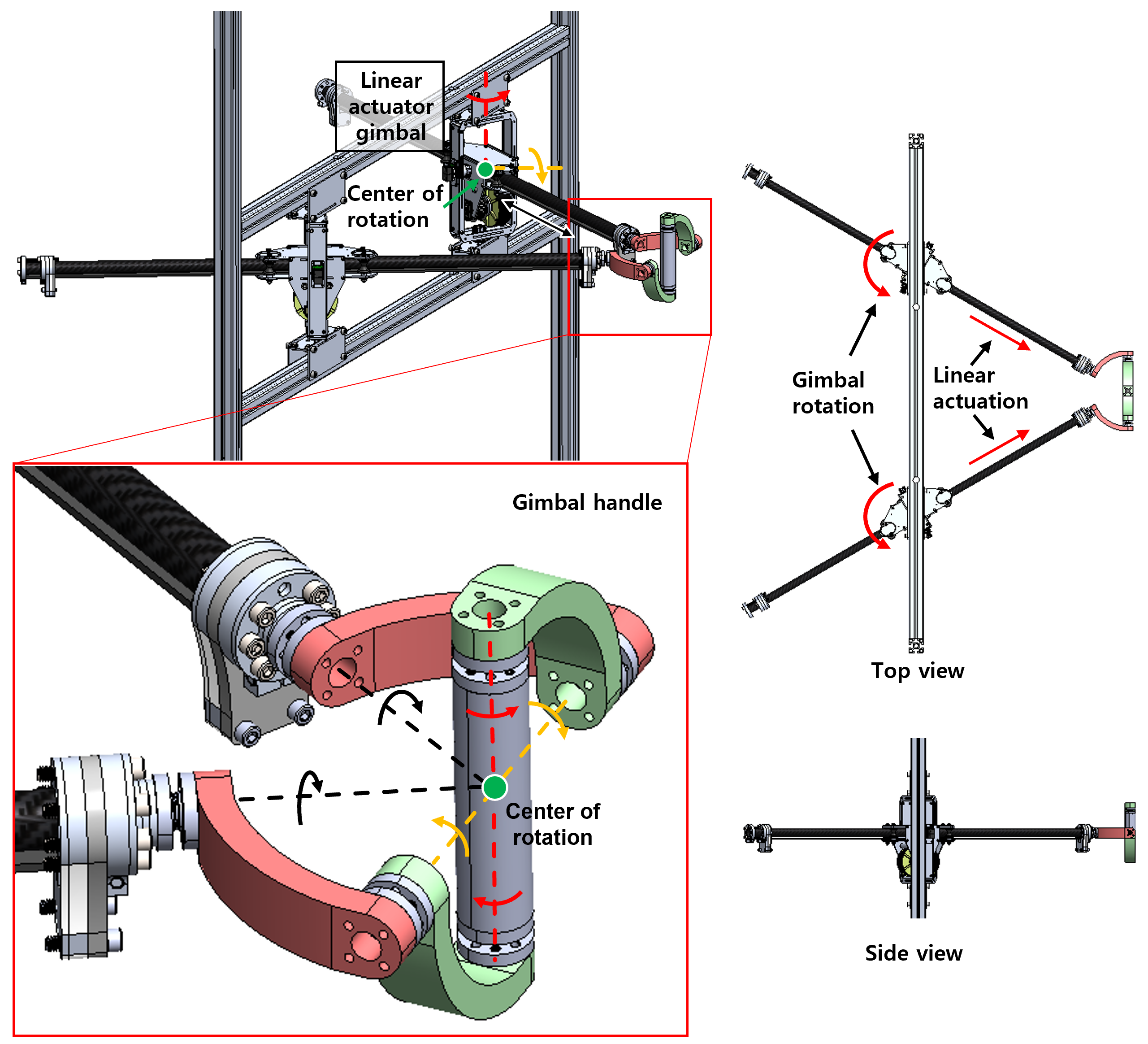}}\\
    \subfloat[Linear motion guidance mechanism of the actuator.]{\label{fig:Flexure}\includegraphics[width = \columnwidth]{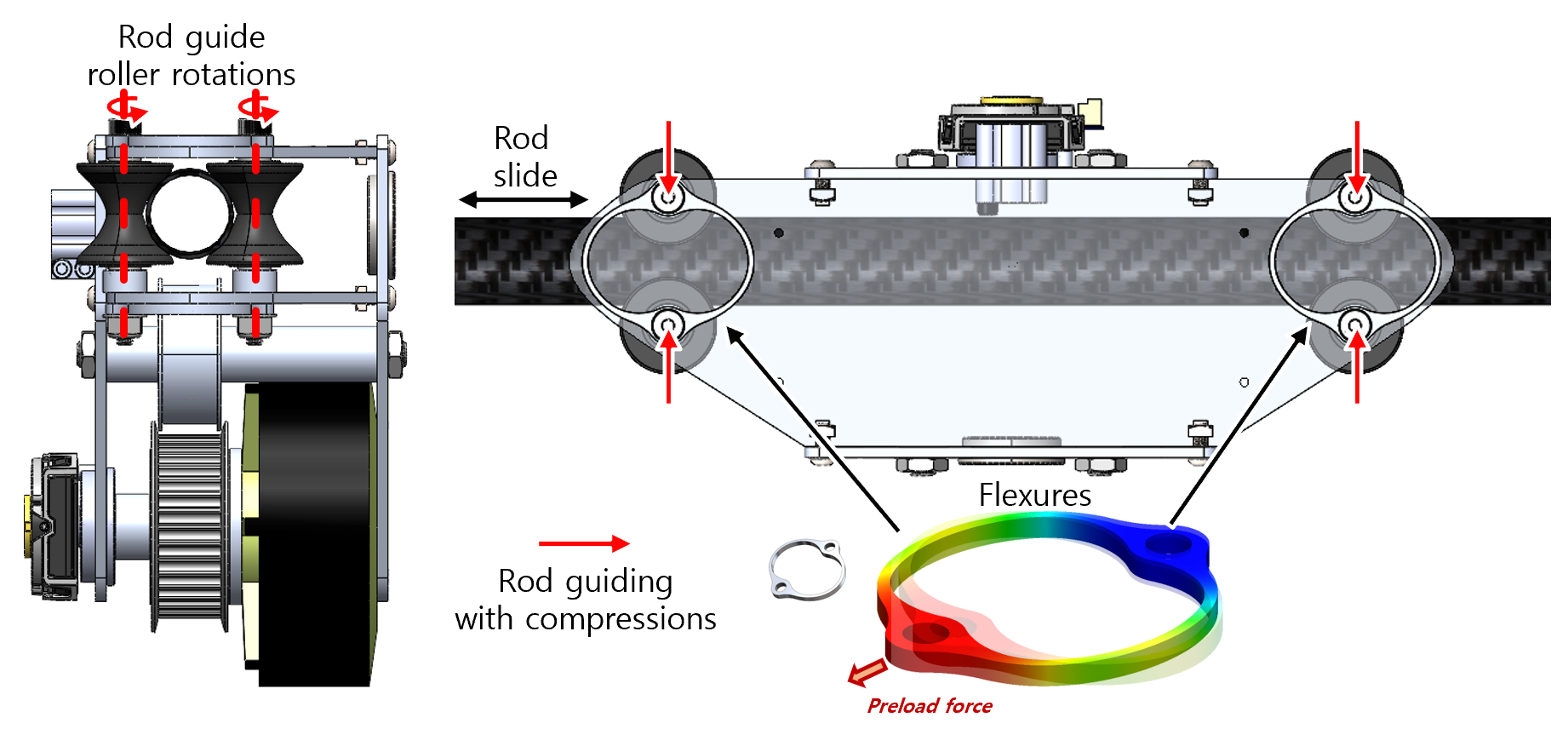}}
    \caption{Detailed design of the proposed haptic interface.}
    \label{fig:Proposed}
\end{figure}

The limitations of previous haptic systems are mostly induced by friction and inertia from gearing \cite{SimICRA21}, and low stiffness caused by bending of the linkages. We propose a haptic interface with modular linear actuators which can overcome those limitations (Figure~\ref{fig:Overview}). The proposed haptic interface has modular linear actuators that were proposed in~\cite{JRNL:Sunyu_HMI}, which are connected to the end-effector in parallel. The actuator has high-torque and low-inertia motors and the timing belt-based reduction system which enable high force capability up to 100 N and backdribable actuation with small friction, resulting in high transparency of the haptic interface. The actuator can be rotated with a 2-DOFs gimbal, which holds the carbon fiber rod with flexures and rollers to constraint the rod and other parts for the rotation while allowing its linear actuation (Figure~\ref{fig:Flexure}). Please refer~\cite{JRNL:Sunyu_HMI} for the detailed actuator design. 

The gimbal handle constraints two linear actuators and allows 3-DOFs free rotation of the user's hand. All the rotational axis of the gimbal handle and the linear actuator axis cross the center of rotation of the gimbal handle. Thus, only the desired forces are delivered to the user's hand without any residual torque. The gimbal handle also allows small deflection of the structure because it has small size compared to the range of linear actuators. The deflection of the linear actuators can be very small because only axial loads are applied to the linear actuators because of the gimbal. The axial stiffness of the linear actuator is dependent on the timing belt stiffness, which is usually very high and has varieties of options. Furthermore, the parallel actuation mechanism of the system can enhance the stiffness at the end-effector. The range of motion of end-effector can be enlarged by increasing the span between the linear actuator gimbal and the linear motion range of the actuator. This does not compromise inertia or friction, as the available force is not changed by the range of motion and the timing belt and linear rod of the actuator have very small inertia. Thus, the system is scalable unlike previously developed systems mentioned in Section I, satisfying stiffness requirements together with a large range of motion and without significantly increasing the inertia of the system. The proposed haptic interface design can be expanded to the device with a larger number of degrees of freedom and feedback force/torque by changing the number of modular linear actuators and applying a suitable end-effector design that can connect the linear actuators. In this paper, a 2-DOFs prototype is presented for comparison with conventional mechanisms. Alternatively, for a 3-DoF haptic interface, a spherical gimbal such as the \textit{Agile Eye} can be implemented \cite{AgileEye}.

\ctable[
caption = Specifications of the 2-DOFs haptic interface.,
label = tab:Spec
]{rccc}{}{\FL &&Specification\ML
&Min. available force&83 N (Continuous)\NN
&at home position&166 N (Peak)\NN
&Motor&Turnigy 9235-100KV\NN
&&Pulley radius: 28.245 mm\NN
&&Nominal torque: 2.825 Nm\NN
&Controller&NI cRIO-9082\NN
&&Control rate: 1 kHz\NN
&Linear motion rod&Material: CFRP\NN
&&OD: 25.4 mm\NN
&&ID: 22.86 mm\NN
&&Length: 1219 mm\NN
&&Weight: 158 $g$\NN
&Timing belt&Gates PowerGrip GT3 5MGT\NN
&&Glass fiber reinforced \NN
&Gimbal handle&Material: ABS and steel\NN
&&Weight: 562 $g$\LL
}

Table~\ref{tab:Spec} shows detailed specifications of the proposed haptic interface. The total linear inertia of the moving parts that includes the rod, timing belt, and belt tensioner is 0.533 kg. The gimbal handle linkages are manufactured by 3D printer, while the joints are composed of metal parts.

\section{Analysis}
\begin{figure}
\centering 
    \includegraphics[width = \columnwidth]{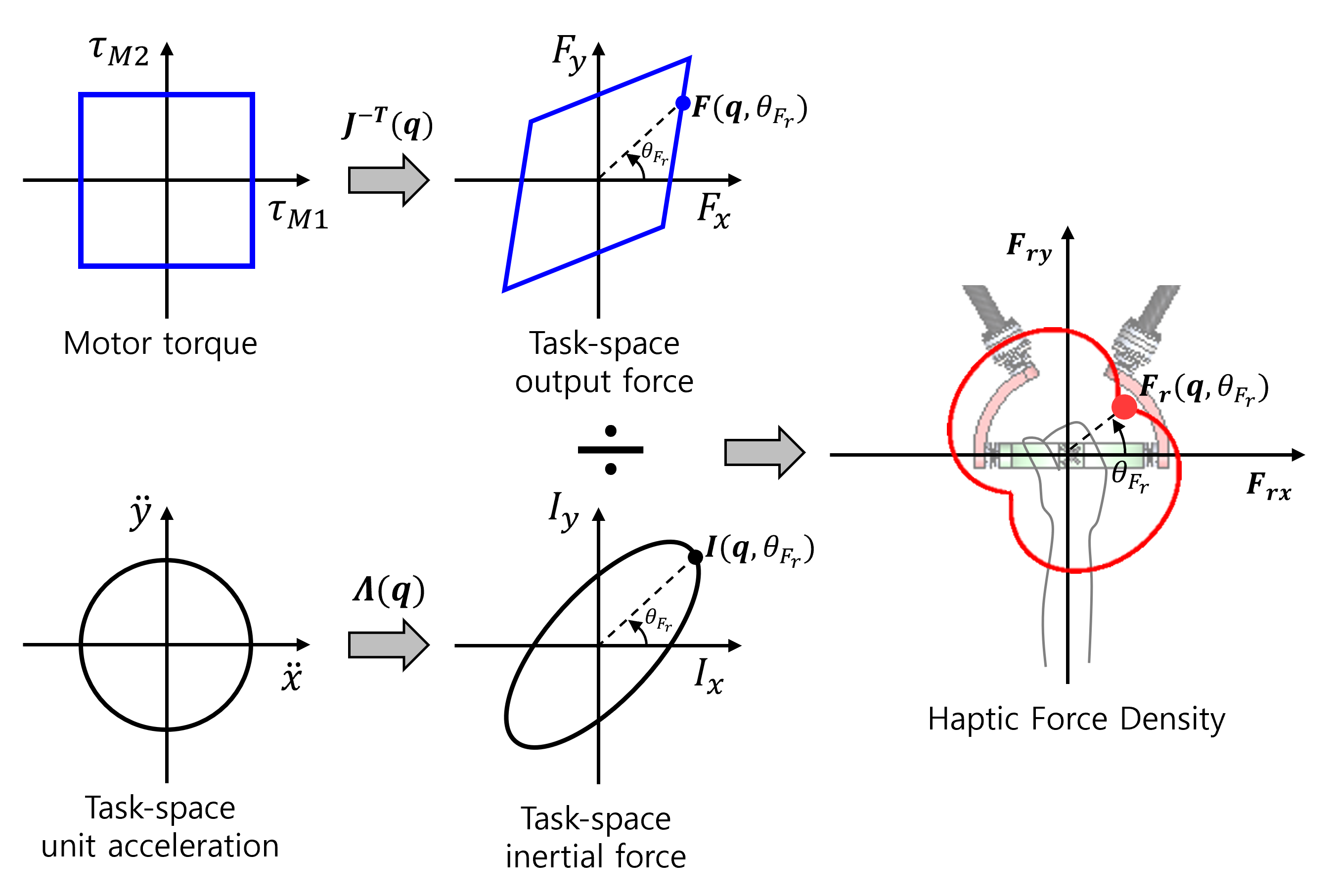}
    \caption{The haptic force density $\boldsymbol{F_{r}}(\boldsymbol{q},\theta_{F_{r}})$ is given by the device's force capability per unit reflected inertia for a certain direction.}
    \label{fig:Performance_index}
\end{figure}

The advantage of the proposed linear actuator-based haptic interface is analyzed in this section, compared with conventional rotational mechanisms. A key metric we can use in this analysis is reflected inertia, because high-frequency performance of the haptic interface is dominated by inertia~\cite{CONF:DM2}. On the other hand, the force capability of the device is also considered as an important metric, as well as the reflected inertia. In this paper, we introduce the concept of \textbf{\textit{Haptic Force Density}} (Fig.~\ref{fig:Performance_index}) which is the force capability per unit reflected inertia as follows: 
\begin{equation}\label{eqn:index}
   \boldsymbol{F_{r}}(\boldsymbol{q},\theta_{F_{r}})=\boldsymbol{u}\frac{|\boldsymbol{F}(\boldsymbol{q},\theta_{F_{r}})|}{|\boldsymbol{I}(\boldsymbol{q},\theta_{F_{r}})|},
\end{equation}
where $\boldsymbol{u}$ represents the unit vector that corresponds to the angle $\theta_{F_{r}}$, $\boldsymbol{q}$ is the actuator displacement vector, $\boldsymbol{F}(\boldsymbol{q},\theta_{F_{r}})$ is the force that the device can generate at the end-effector in a quasi-static state, which can be calculated as follows:
\begin{equation}\label{eqn:force}
   \boldsymbol{F}(\boldsymbol{q},\theta_{F_{r}})=\boldsymbol{J^{-T}}(\boldsymbol{q})\boldsymbol{\tau}
\end{equation}
where $\boldsymbol{J}$ is the Jacobian. $\boldsymbol{I}(\boldsymbol{q},\theta_{F_{r}})$ is the reflected inertia perceived by the user when the end-effector is back-driven with a unit acceleration (1m/$s^{2}$), and can be calculated as follows~\cite{BOOK:Modern_Robotics}:
\begin{equation}\label{eqn:inertia}
   \boldsymbol{I}(\boldsymbol{q},\theta_{F_{r}})=\boldsymbol{J^{-T}}(\boldsymbol{q})\boldsymbol{M}\boldsymbol{J^{-1}}(\boldsymbol{q})\boldsymbol{\ddot{x}}=\boldsymbol{\Lambda}(\boldsymbol{q})\boldsymbol{\ddot{x}}
\end{equation}
Here, $\theta_{F_{r}}$ represents the direction of the generated force that corresponds to a set of actuator torques $\boldsymbol{\tau}$ or the back-driven task-space inertial force direction, and $\boldsymbol{M}$ represents the inertia matrix which is determined by the actuator inertia and the mechanism of the system. 

Higher $\boldsymbol{F_{r}}(\boldsymbol{q},\theta_{F_{r}})$ means the displayable force magnitude to a specific direction $\boldsymbol{u}$ that corresponds to the angle $\theta_{F_{r}}$ is larger compared to the magnitude of reflected inertia when the haptic device is back-driven by the user to the same direction. Since $\boldsymbol{F_{r}}(\boldsymbol{q},\theta_{F_{r}})$ is related to the Jacobian, which is determined by the geometry of the device, its characteristics are varied by the actuation mechanism.

\begin{figure}
    \centering
    \subfloat[Parallel linear mechanism (proposed).] {\label{fig:Serial_Linear}\includegraphics[width = 3.2in]{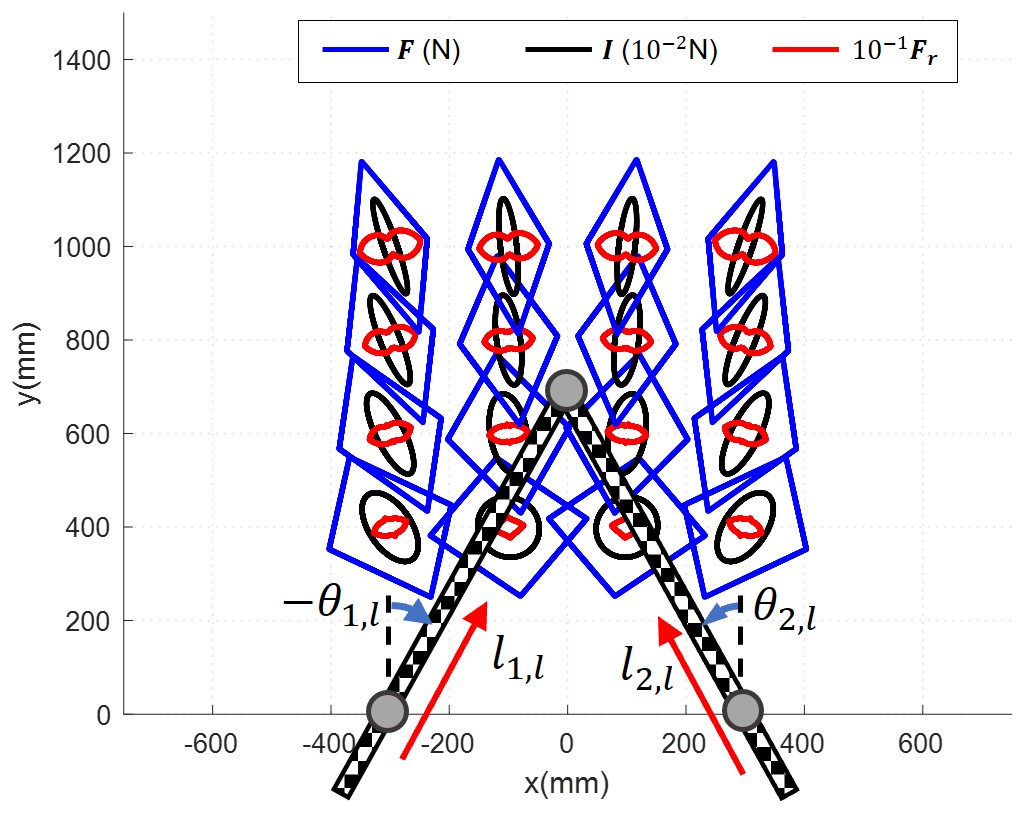}}\\
    \subfloat[Serial rotational mechanism.]{\label{fig:Serial_Rotational}\includegraphics[width = 3.2in]{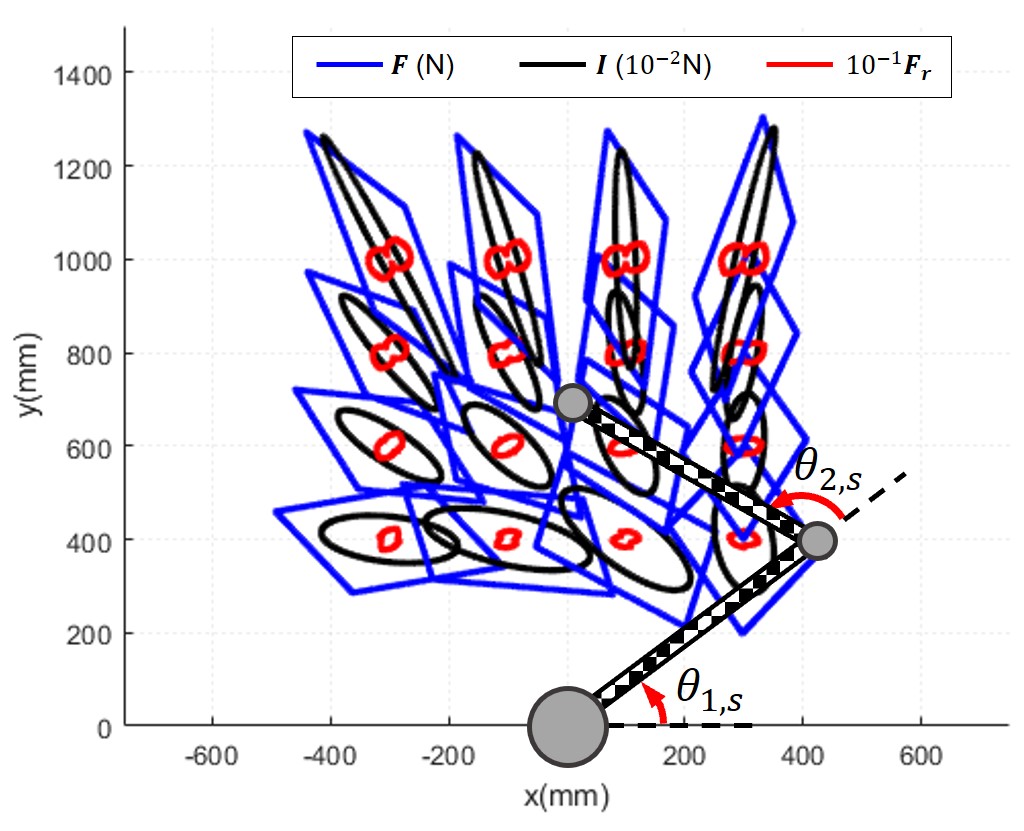}}\\
    \subfloat[Parallel rotational mechanism.]{\label{fig:Parallel_Rotational}\includegraphics[width = 3.2in]{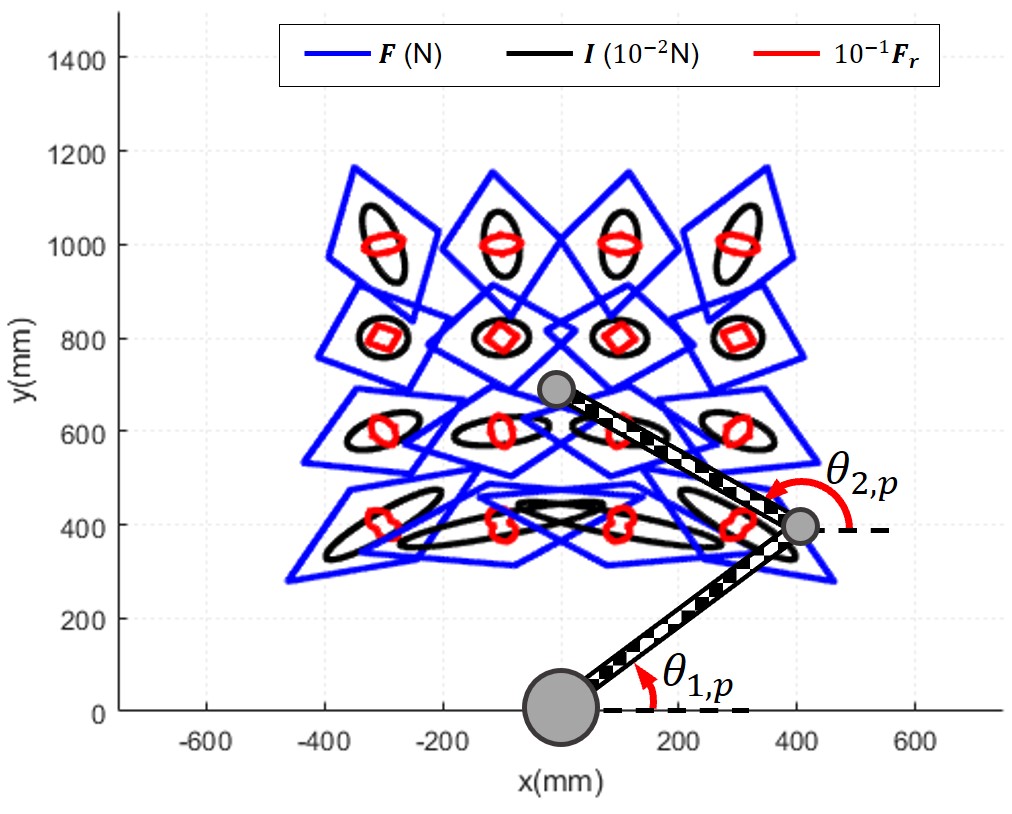}}
    \caption{Force capability, inertial force and their ratio of haptic interface mechanisms.}
    \label{fig:Fore_analysis}
\end{figure}

Figure~\ref{fig:Fore_analysis} shows the force capability, reflected inertial force, and their ratio of a haptic interface with (a) parallel linear mechanism (this work), (b) serial rotational mechanism, (c) parallel rotational mechanism. The motors are assumed to be placed at the base parts, since this can minimize the inertia induced by the motor mass. The parallel linear mechanism represents the proposed haptic interface mechanism in this paper. This mechanism consists two passive rotational joints ($\theta_{1,l}$ and $\theta_{2,l}$) and two linear joints that is actuated by the motor ($l_{1,l}$ and $l_{2,l}$). The other two mechanisms represent conventional haptic interface designs, which utilize the rotational actuation mechanism. In the serial rotational mechanism~\cite{JRNL:L-EXOS,CONF:DLRHaptics,JRNL:HIROIII}, the first motor actuates the first joint that corresponds to $\theta_{1,s}$, and the second motor is attached to the first link. The second motor rotates the second joint ($\theta_{2,s}$) relatively to the first linkage. On the other hand, the motors of the parallel rotational mechanism are independent. Both motors are fixed to the base structure, and connected to the two joints independently. As the result, the joint motion is only dependent on the motion of each motor. The parallel rotational mechanism is considered as a linearized version of linkage-based mechanisms in~\cite{WEB:Delta,WEB:Phantom,WEB:HD2,WEB:Cyberforce,JRNL:VirtuaPower,CONF:Mantis}, assuming the transmission for the second joint is rigid enough.

The linkage lengths of rotational mechanisms were set to 600 mm each to cover the required range of motion (600 mm x 600 mm) for a quadrant of a human, from the chest to the maximum extension of the arm, without significant singularity in the workspace. The same motor of the proposed linear actuator-based haptic interface was assumed. 21:1 and 18:1 of gear ratios were applied for the serial and parallel rotational mechanism, respectively. These were selected to generate the same minimum force of the parallel linear mechanism at the home position shown in the figure. 

The blue lines shown in Fig.~\ref{fig:Fore_analysis} are the force that the mechanisms can generate at each position for all directions ($\boldsymbol{F}(\boldsymbol{q},\theta_{F_{r}}))$, and the black ellipsoids show the reflected inertia of the mechanisms when the end-effector is back-driven ($\boldsymbol{I}(\boldsymbol{q},\theta_{F_{r}})$). The haptic force density $\boldsymbol{F_{r}}(\boldsymbol{q},\theta_{F_{r}})$ was calculated by dividing the force by the inertia for all directions. As shown in the figures, the force and inertia show different characteristics by the mechanisms. The parallel linear mechanism and the parallel rotational mechanism show cartesian coordinate-based characteristics, while that of the serial rotational mechanism seems dependent on the polar coordinate.

\begin{figure}
\centering 
    \includegraphics[width = \columnwidth]{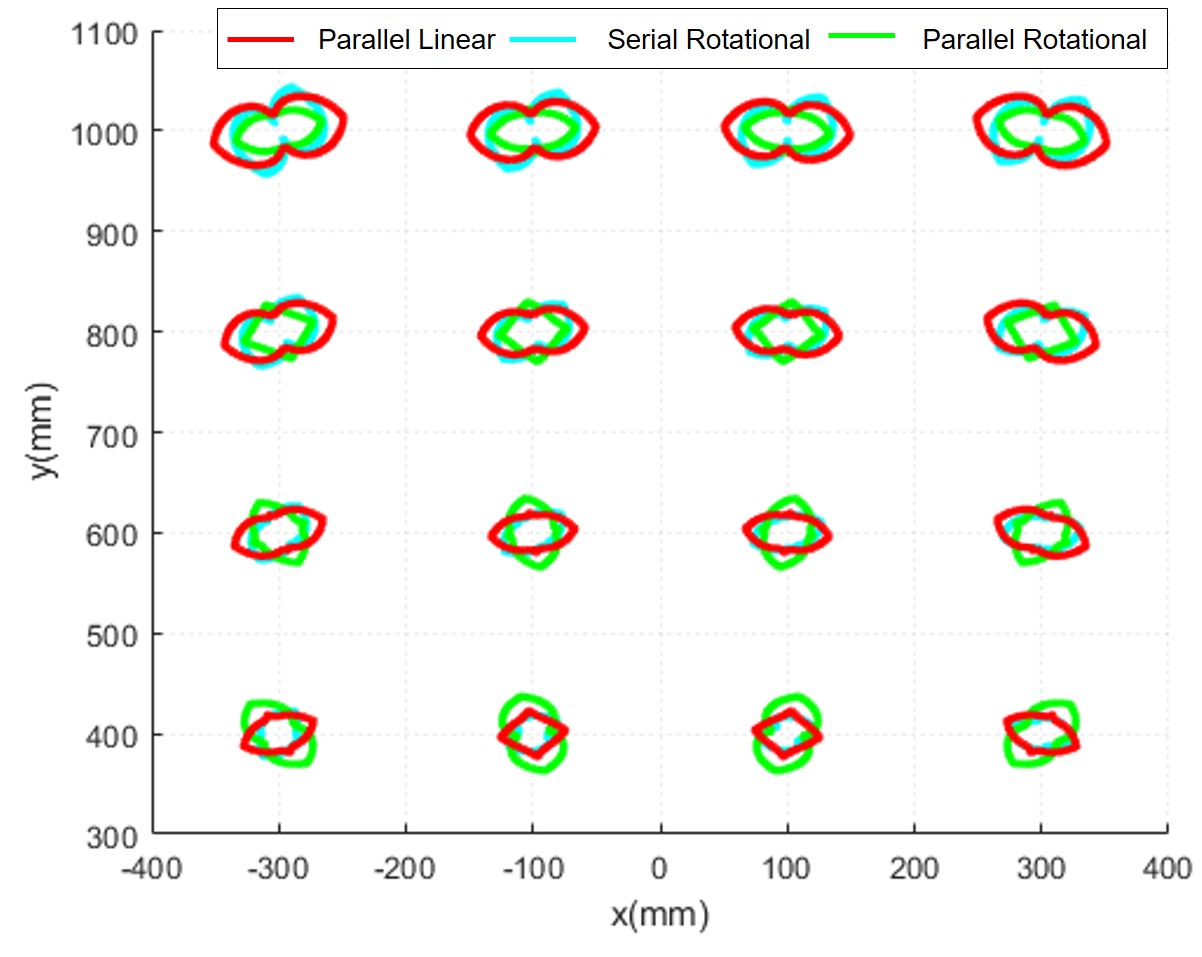}
    \caption{Haptic force density of the parallel linear mechanism and the rotational mechanisms. The values are scaled down to 1/10.}
    \label{fig:Force_ratio}
\end{figure}

Figure~\ref{fig:Force_ratio} shows overlapped $\boldsymbol{F_{r}}(\boldsymbol{q},\theta_{F_{r}})$ of the three mechanisms. As shown in the figure, the parallel linear mechanism shows the largest haptic force density for most of the cases. The minimum $\boldsymbol{F_{r}}(\boldsymbol{q},\theta_{F_{r}})$ for each mechanism was 0.875, 1.299 and 1.706 for the serial rotational, parallel rotational and parallel linear mechanisms, respectively. This implies that the parallel linear mechanism can ensure better haptic experiences in terms of force capability over the reflected inertia. However, the difference is not overwhelming; this can be changed by detailed design.

\begin{figure}
\centering 
    \includegraphics[width = \columnwidth]{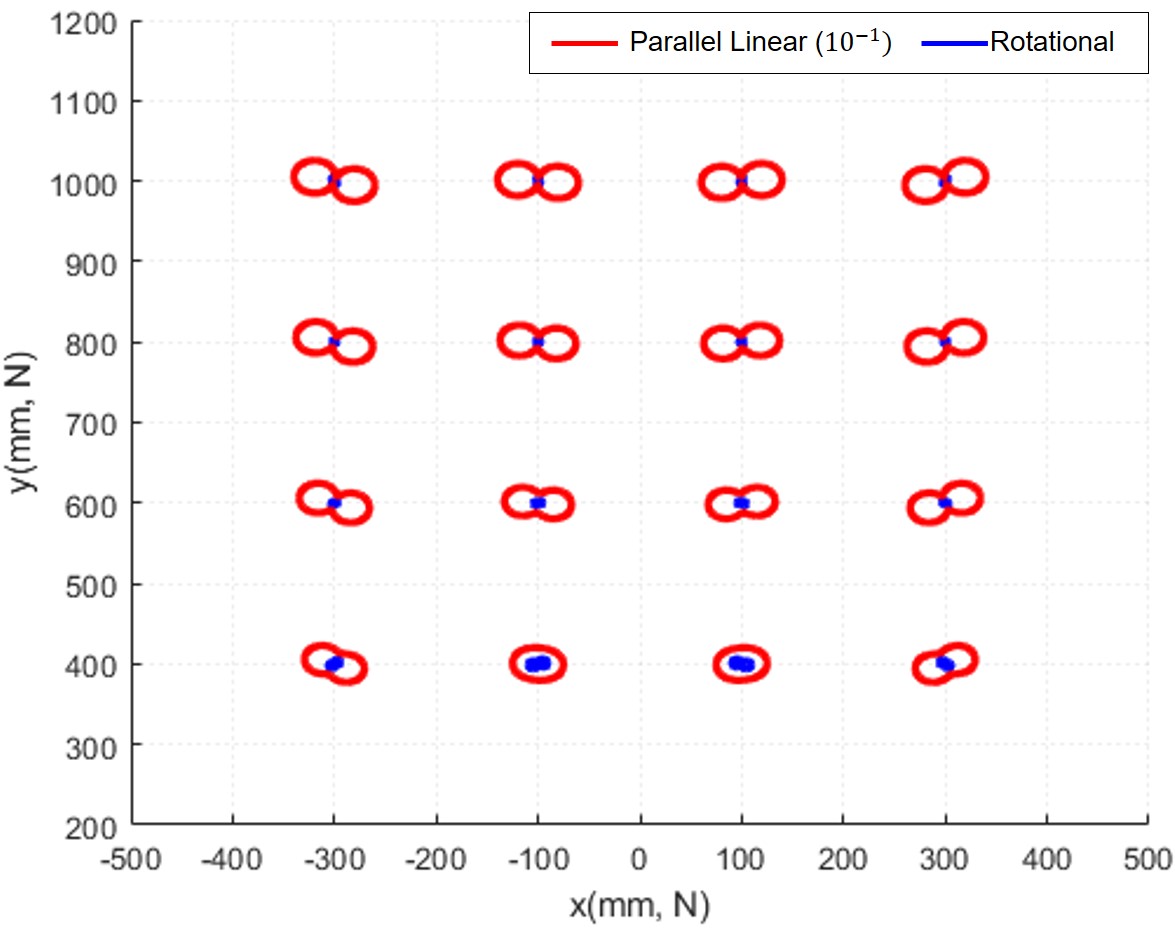}
    \caption{Structural stiffness of the rotational mechanisms and the parallel linear mechanism.}
    \label{fig:Stiffness_structural}
\end{figure}

The other advantage of the proposed haptic interface mechanism is the structural stiffness. As the force is delivered to the end-effector with a linear axis, there is no bending deflection for the linear actuator. Since the linear stiffness is determined by the timing belt stiffness, the resulting end-effector stiffness can be as stiff as the timing belt. Figure~\ref{fig:Stiffness_structural} shows the structural stiffness of the linear parallel and rotational mechanisms. In this analysis, both rotational actuators and parallel linear actuators are fixed at several target points. The rods of the rotational mechanisms were considered to have the same rod dimension and properties of the carbon fiber rod applied for the proposed 2-DOFs haptic interface. Then, 1 mm of displacement was applied to the end-effector for all directions. The blue and red lines represent the resulted force by the displacement, which also can be considered as the structural stiffness of the mechanism in the unit of N/mm. Note that the force of the parallel linear mechanism was scaled to 1/10 for the visualization. As shown in Fig.~\ref{fig:Stiffness_structural}, the structural stiffness of the parallel linear mechanism is significantly larger in all configurations. The minimum stiffness of the rotational mechanism was 1.29 N/mm, while that of the linear parallel mechanism was 47.6 N/mm. 


The stiffness of the rotational mechanism is determined by the radial stiffness of the rod, which is inversely proportional to the cube of the length of the rod. On the other hand, the displacement of the parallel linear mechanism comes from the linear deformation of the timing belt, which has very high stiffness and is inversely proportional to the length. \textit{Thus, the linear parallel mechanism can enlarge the ROM without sacrificing the inertia, while the rotational mechanisms should increase the inertia to satisfy the required stiffness}. The stiffness of rotational mechanisms can be enhanced by introducing parallel linkage structures, like a delta mechanism and the five-bar linkage mechanism introduced in many haptic interfaces introduced in Section I. However, that can only linearly enhance the stiffness by the number of the parallel link sets, while the stiffness is more aggressively reduced by the inverse cube of the rod length. This stiffness analysis results imply that the proposed parallel linear mechanism is \textit{scalable}, while conventional rotational mechanisms are not.

\section{Control}

The linear actuator dominates the haptic interface performance because they are connected to a gimbal handle that has a small dimension and thus can present small mass and deformation. In this section, the inertia, force, friction, and resulting force bandwidth and available stiffness of the actuator are identified for the verification of the actuator performance.

\begin{figure}
\centering 
    \includegraphics[width = 1.8in]{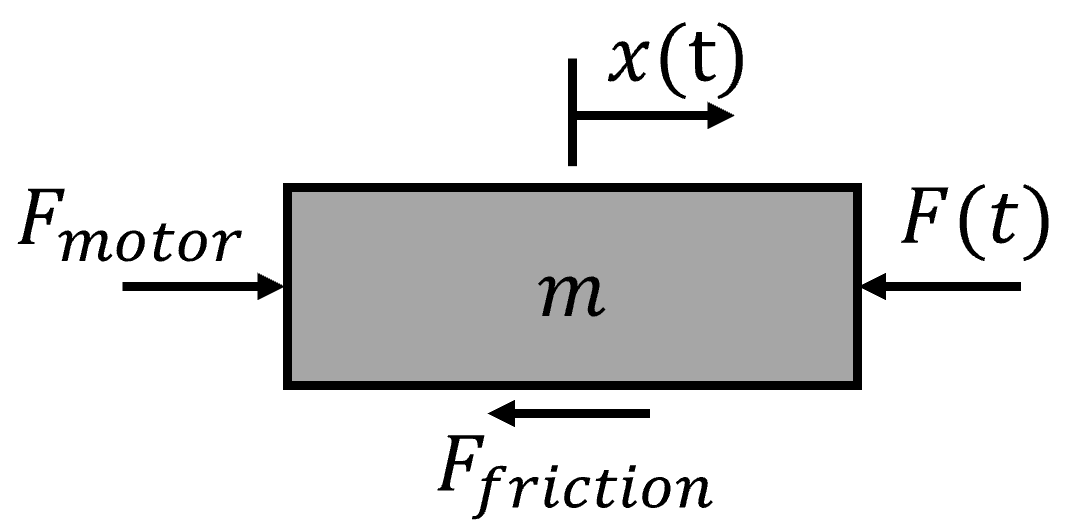}
    \caption{The dynamic model of the linear actuator.}
    \label{fig:Actuator_Model}
\end{figure}

The actuator was modeled as a linear system with lumped inertia composed of the rotor and rod, in addition to Coulomb friction and viscous friction from the bearings (Fig.~\ref{fig:Actuator_Model}). Motor torque was assumed to be linearly proportional to the applied current. Then, the output force $F(t)$ of the actuator that is applied to the environment can be obtained from the equation of motion as follows:
\begin{equation}\label{eqn:sys_model1}
   F(t)=-F_{inertia}-F_{friction}+F_{motor}
\end{equation}
where
\begin{eqnarray}
   F_{inertia}&=&m\ddot{x}(t)\label{eqn:sys_model2}\\
   F_{friction}&=&F_{static}sign(\dot{x}(t))+a_{viscous}\dot{x}(t)\label{eqn:sys_model3}\\
   F_{motor}&=&K_{m}u(t)\label{eqn:sys_model4}
\end{eqnarray}
Here, $x$ is the linear travel distance measured by the motor encoder and transmission geometry, $m$ is the reflected inertia of all actuated parts of the linear actuator, $F_{static}$ is the static friction constant, $a_{viscous}$ is the viscous friction constant, $K_{m}$ is the motor constant, and $u$ is the control input for the motor driver.

\begin{figure}
    \centering
    \subfloat[System identification result for the inertia and friction.]{\label{fig:SystemId}\includegraphics[width = \columnwidth]{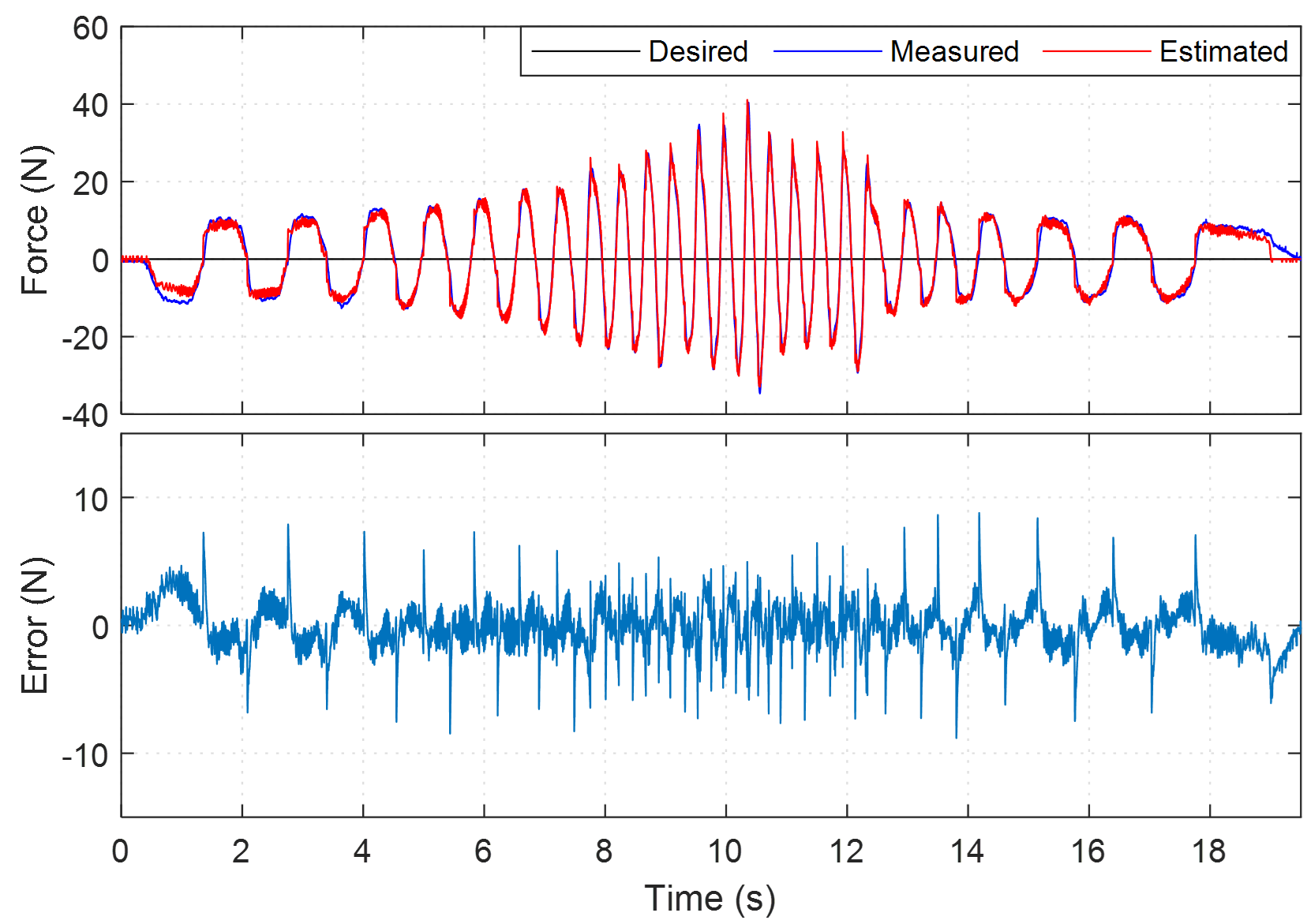}}\\
    \subfloat[System identification result for the motor constant.]{\label{fig:SystemId_Motor}\includegraphics[width = \columnwidth]{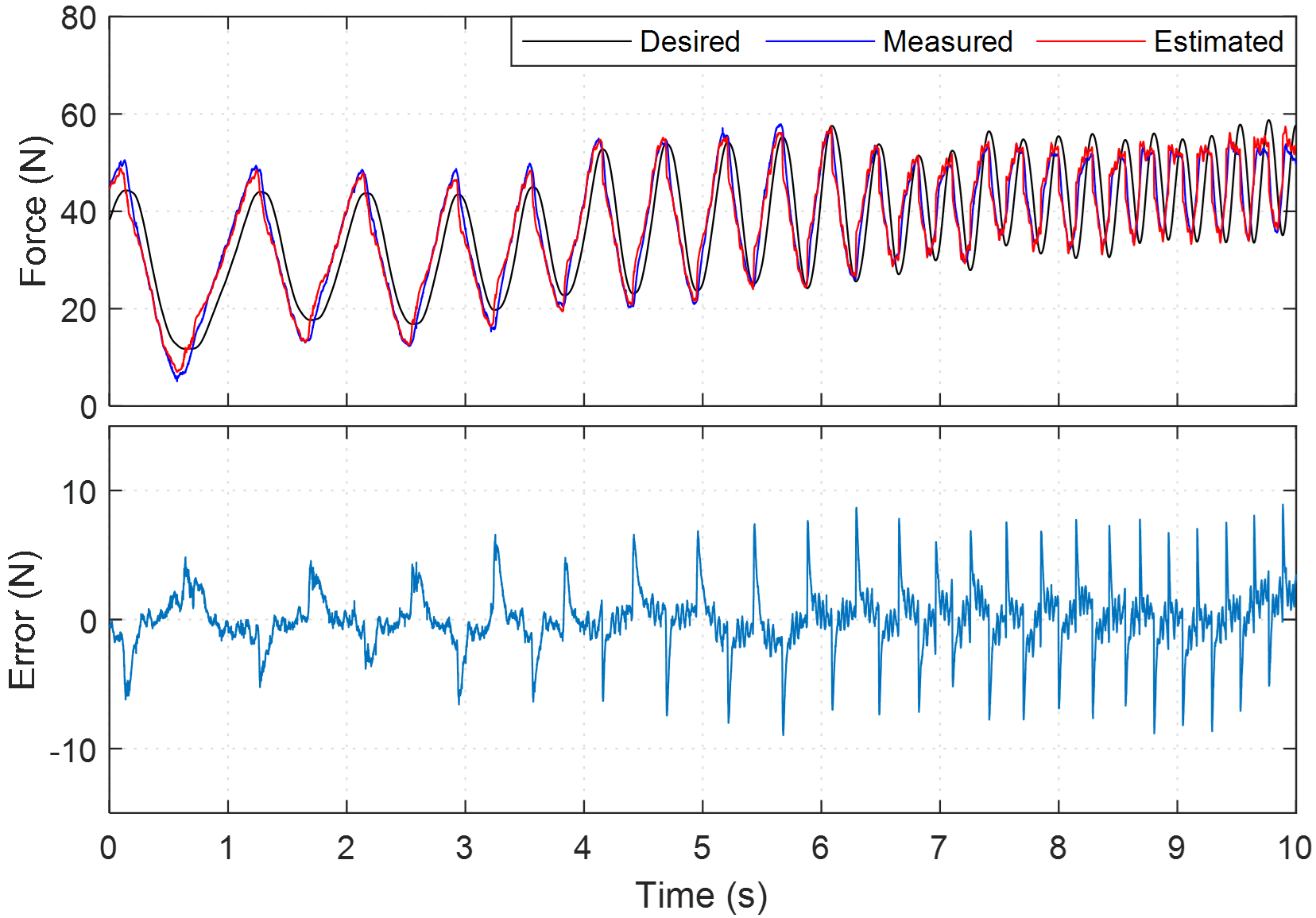}}
    \caption{System identification of the linear actuator.}
    \label{fig:SysID}
\end{figure}

A zero current was applied to the motor to identify the reflected inertia and the friction. A human applied a chirp force to the linear actuator by holding a load cell mounted on the end-effector of the actuator, while the displacement and load cell force were recorded. A least-square method was used using the measured force from the load cell. Velocity and acceleration were calculated using the time-derivative of $x(t)$, measured from the motor encoder. The University of Illinois at Urbana-Champaign Institutional Review Board has reviewed and approved this research study, as well as other research studies in Section IV and V. The identified values are $m=1.116$ kg, $F_{static}=5.26$ N, and $a_{viscous}=7.52$ Ns/m. Fig.~\ref{fig:SystemId} shows the desired, measured, and estimated force with the identified model, and the estimation error. The root mean square (RMS) error and maximum magnitude error were 1.90 N and 9.98 N, respectively. The rotor and idler pulley reflected inertia were estimated as 0.583 kg by subtracting the inertia of the rod. The identified friction and inertial parameters indicate that this actuator is highly backdrivable and has very small inertia. As discussed in section II, these characteristics do not change that much even if the range of motion is expanded from the current 1.1 m, since the carbon fiber rod and timing belt inertia are small. 


\begin{figure}
\centering 
    \includegraphics[width = \columnwidth]{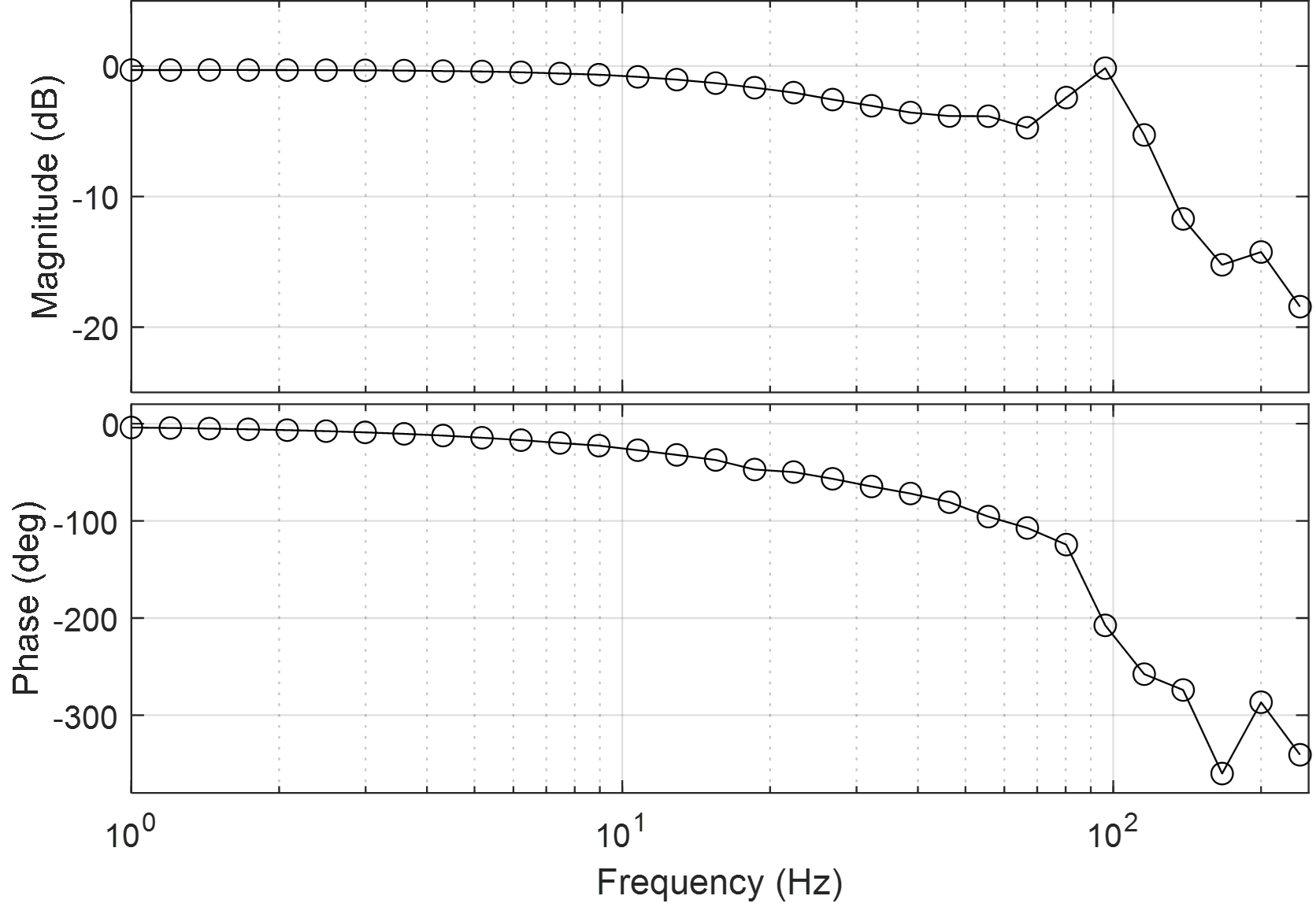}
    \caption{Frequency response of the linear actuator.}
    \label{fig:Freq_Response}
\end{figure}

\begin{figure}
\centering 
    \includegraphics[width = \columnwidth]{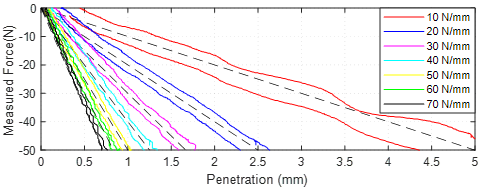}
    \caption{Displayable stiffness of the linear actuator. Dotted line is the desired stiffness while solid line represents experimental data.}
    \label{fig:Stiffness}
\end{figure}

\begin{figure}
\centering 
    \includegraphics[width = \columnwidth]{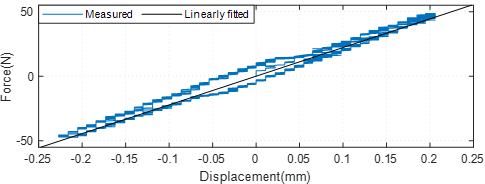}
    \caption{Timing belt stiffness identification.}
    \label{fig:Timing_Stiffnes}
\end{figure}

After identifying the parameters that determine the natural response, the motor constant was identified in a similar manner. A soft virtual spring force with an arbitrary stiffness was applied to the actuator, and a human moved the end-effector with chirp force. The identified motor constant was 3.25 N/A. Figure~\ref{fig:SystemId_Motor} shows the verification of the identified motor constant in a similar experiment. The desired force was generated with a soft virtual spring of stiffness 0.4 N/mm, and the actuator end-effector force was measured by the load cell. The RMS and maximum estimation errors were 2.19 N and 8.96 N, respectively. Note that the actuator can generate up to 100 N~\cite{JRNL:Sunyu_HMI} and requires no feedback from the load cell.

The frequency response of the linear actuator was tested to show its force capability in different frequencies. The end-effector was fixed to a solid frame, and sinusoidal reference force was commanded to the actuator with different frequencies. The magnitude was set to 50 N, which is half of the maximum force of the linear actuator. The results (Fig.~\ref{fig:Freq_Response}) show that the actuator force response has a cutoff frequency near 30 Hz, and shows resonance around 100 Hz. Considering the characteristics of parallel linear actuation mechanism, the haptic interface can show similar force bandwidth if the end-effector is lightweight and the connection is stiff enough. Also, the longitudinal stiffness of the timing belt was identified to 222 N/mm by fitting a linear spring model to the measured data at low frequency (1 Hz, Fig.~\ref{fig:Timing_Stiffnes}). Note that the actuator stiffness can be further reinforced by selecting a timing belt with higher longitudinal stiffness for the extended range of motion.

To investigate the maximum stiffness of the linear actuator that can be provided to the user, we conducted a stiffness rendering experiment. A human slowly pulled and released the load cell on the end-effector of the linear actuator while a virtual wall was rendered varying its stiffness at each trial. Figure~\ref{fig:Stiffness} shows the experimental result. As shown in the figure, the actuator can provide linear force feedback up to 70 N/mm of stiffness. Rendering a virtual wall stiffer than 70 N/mm showed unstable interaction with the user when the user was releasing the actuator. This might be caused by the quantization of the position by the digital encoder of 12 bits.

However, 70 N/mm of stiffness is sufficiently high to render a stiff wall, considering the maximum virtual stiffness of the commercial haptic devices are 3 to 12 N/mm within a small range of motions. Since parallel mechanisms enhance the stiffness with the structural reinforcement, the end-effector stiffness can be higher if a high stiffness handle is provided. 

The end-effector force of the haptic interface was controlled by a simple feedforward (open-loop) force control using Jacobian assuming quasi-static motion. The required joint-space torque $\boldsymbol{\tau}$ was calculated from the desired task-space force $\boldsymbol{F_{d}}$ as follows:
\begin{equation}\label{eqn:control}
   \boldsymbol{\tau}=\boldsymbol{J^{T}}\boldsymbol{F_{d}}
\end{equation}
The inertia, friction, and other nonlinearities were neglected since the actuator provides high backdrivability as shown in the system identification result. It was not necessary to care about the collision avoidance with the user in the workspace because the linear actuators are extended from the front of the user. Since the system is controlled by a feedforward controller without any force feedback sensor, the system can be stable even in very fast motion and instantaneous contact with a real or virtual object.

\begin{figure}
\centering 
    \includegraphics[width = \columnwidth]{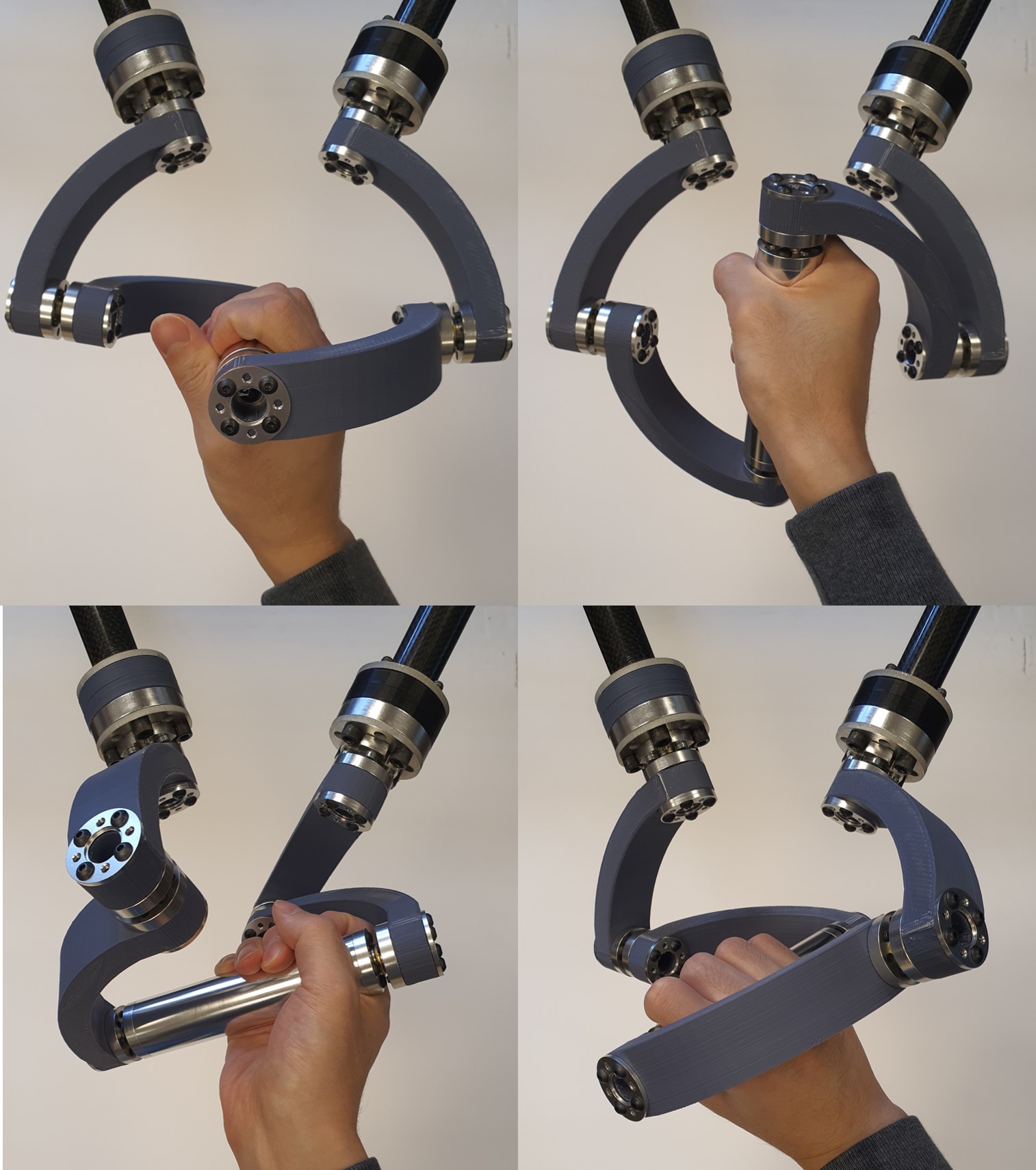}
    \caption{Free rotation with the gimbal handle.}
    \label{fig:Gimbla_Handle_Exp}
\end{figure}

\begin{figure}
\centering 
    \includegraphics[width = \columnwidth]{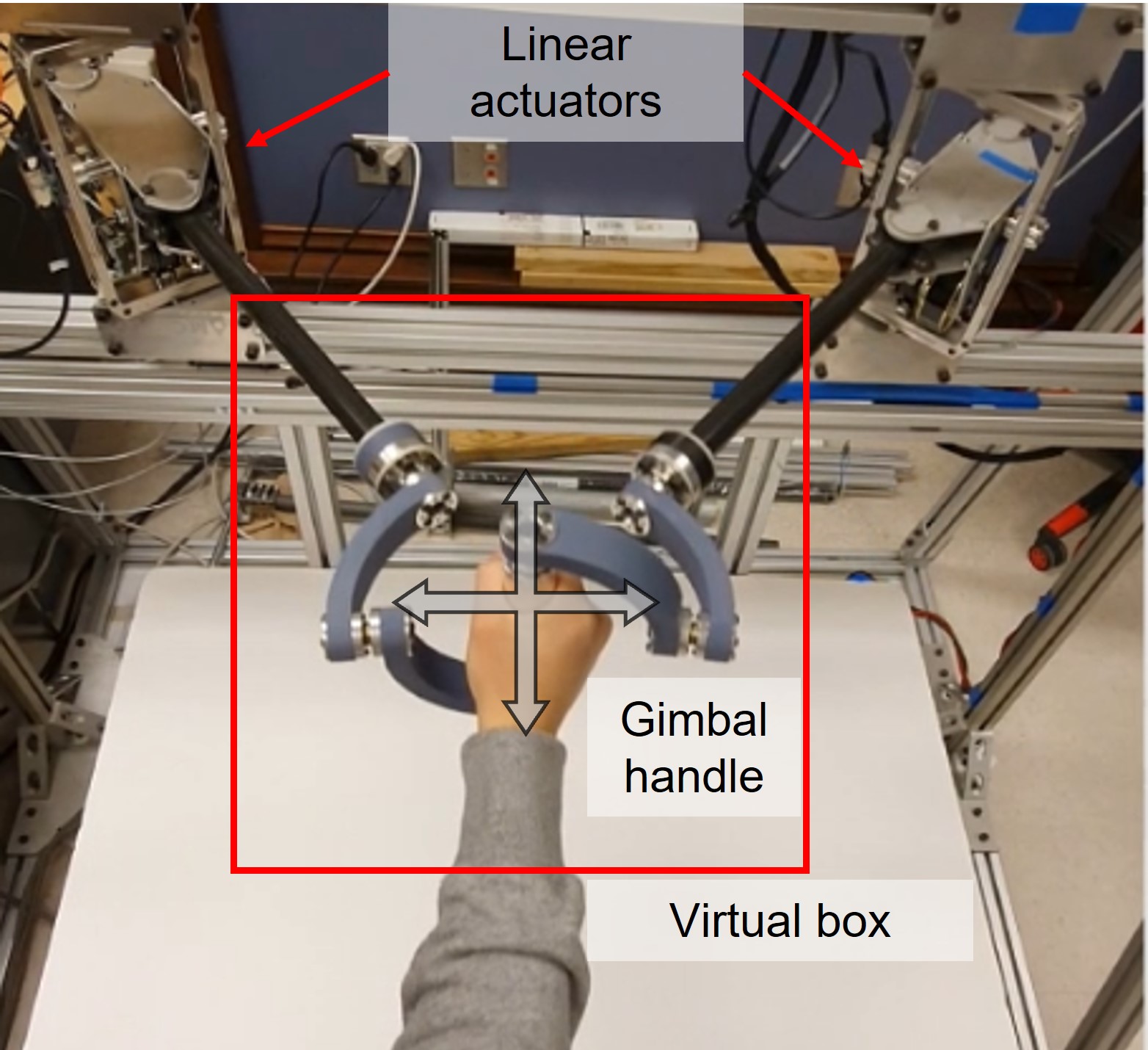}
    \caption{Experimental setup for the interaction with a virtual box test.}
    \label{fig:Virtual_Box_Setup}
\end{figure}

\begin{figure}
    \centering
    \subfloat[Isometric view] {\label{fig:Virtual_Box_iso}\includegraphics[width = \columnwidth]{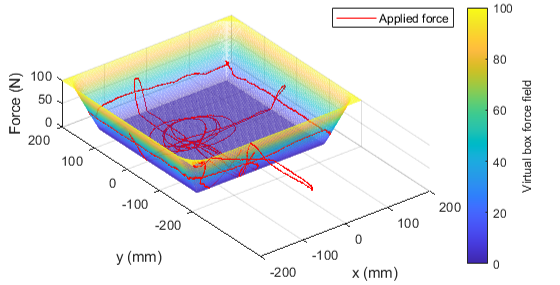}}\\
    \subfloat[Top view and side views.]{\label{fig:Virtual_Box_topside}\includegraphics[width = \columnwidth]{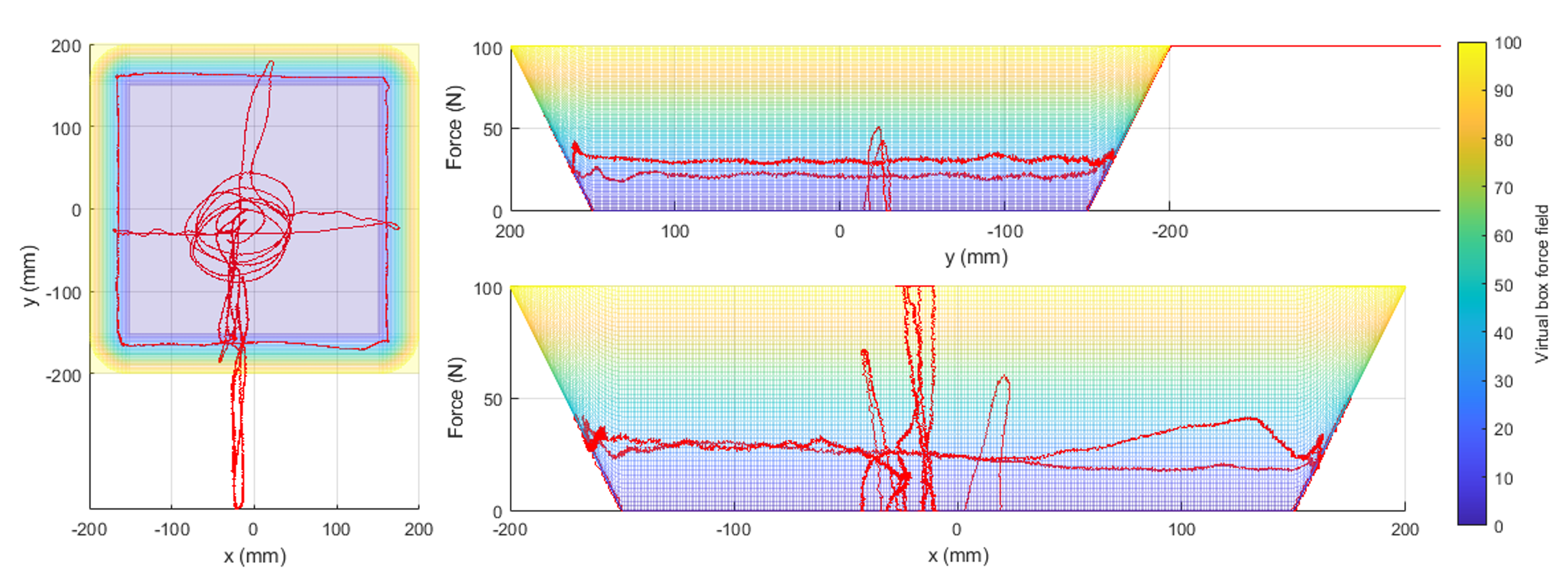}}
    \caption{Virtual box rendered as a potential field. Position and applied force on the end-effector in the interaction with a virtual box test.}
    \label{fig:Virtual_Box}
\end{figure}

\section{Experiment}
The performance of the proposed haptic interface was tested with experiments. The user held the gimbal handle and rotated in several orientations. As shown in Fig.~\ref{fig:Gimbla_Handle_Exp}, the gimbal handle allows free rotation to the user. The available ROMs are about $-45^{\circ}$ to $135^{\circ}$ in pitch, and $-90^{\circ}$ to $-135^{\circ}$ in roll. The yaw motion is allowed in between $-60^{\circ}$ to $60^{\circ}$ until the user's forearm reach to the gimbal. The force delivery of the interface was tested by rendering a virtual box to the user. Figure~\ref{fig:Virtual_Box_Setup} shows the experimental setup. A two-dimensional virtual box with the size of 300 mm x 300 mm was designed, and a virtual force field with 2 N/mm stiffness was set to the edge of the virtual box. The maximum force that could be applied by the actuator was limited to 65 N, while the task-space force was limited to 100 N by the magnitude for safety. The user tested the backdrivability in the free space by moving the end-effector, and interacted with the virtual box with the force feedback. The attached video shows the experiments with the virtual box. As shown in the video, the user can freely rotate the hand by the DOFs of the gimbal handle. It was possible to stably interact with the virtual wall with the guidance of the force field. Figure~\ref{fig:Virtual_Box} shows the result of a similar experiment. The force field of the virtual box is shown as the gradient color. The 2D end-effector trajectory and the force applied at the points are shown as the red line. The user could slide on the virtual wall surface with a contact force of near 25 N without significant effort.

\section{Discussions and Conclusion}
In this paper, we presented a new haptic interface mechanism with linear actuators. The force, inertia, and stiffness analysis showed that the proposed mechanism has advantages over conventional haptic interface mechanisms with the lightweight and high-stiffness parallel structure. A 2-DOFs haptic interface was presented as an example of the parallel linear mechanism haptic interface. The actuators were highly backdrivable without any feedback, while they can provide very high force and stiffness when compared to existing devices. The timing belt transmission also eliminates the use of a high-quality gearbox, which reduces costs and simplifies the reduction ratio design by changing the pulley radius. The experimental results showed that the proposed haptic interface mechanism could allow free rotation of human hands with the gimbal handle, while providing stable interaction with the virtual environment. Since 3D printed plastic parts were used for the gimbal handle and the displayable stiffness of the haptic interface is limited by the structural stiffness, the available stiffness was limited to 2 N/mm. However, this could be improved if stiffer material and design are applied to the handle. The analysis and experiments with the prototype haptic interface imply that the proposed mechanism has potential as a haptic interface that can provide high force and stiffness with a wide range of motion. The scalability and modular design with the linear actuator enable expanding ROM and DOF without sacrificing performance. We are developing 3D force feedback haptic interface with three linear actuators and a new gimbal handle design, and will be discussed in future research.

\bibliographystyle{IEEEtran}
\bibliography{IEEEabrv,YT_bib}

\begin{thebibliography}{10}
\providecommand{\url}[1]{#1}
\csname url@rmstyle\endcsname
\providecommand{\newblock}{\relax}
\providecommand{\bibinfo}[2]{#2}
\providecommand\BIBentrySTDinterwordspacing{\spaceskip=0pt\relax}
\providecommand\BIBentryALTinterwordstretchfactor{4}
\providecommand\BIBentryALTinterwordspacing{\spaceskip=\fontdimen2\font plus
\BIBentryALTinterwordstretchfactor\fontdimen3\font minus
  \fontdimen4\font\relax}
\providecommand\BIBforeignlanguage[2]{{%
\expandafter\ifx\csname l@#1\endcsname\relax
\typeout{** WARNING: IEEEtran.bst: No hyphenation pattern has been}%
\typeout{** loaded for the language `#1'. Using the pattern for}%
\typeout{** the default language instead.}%
\else
\language=\csname l@#1\endcsname
\fi
#2}}

\bibitem{JRNL:Ryu_Haptic_Passivity}
B.~Hannaford and J.~Ryu, ``Time-domain passivity control of haptic
  interfaces.'' \emph{IEEE transactions on Robotics and Automation}, pp. 1--10,
  2002.

\bibitem{JRNL:Khatib_Haptic}
J.~Park and O.~Khatib, ``A haptic teleoperation approach based on contact force
  control,'' \emph{The International Journal of Robotics Research}, pp.
  575--591, 2006.

\bibitem{JRNL:Sirouspour_Haptics_Transparency}
A.~Abdossalami and S.~Sirouspour, ``Adaptive control for improved transparency
  in haptic simulations,'' \emph{IEEE transactions on Haptics}, pp. 2--14,
  2009.

\bibitem{JRNL:Pan_Haptics_Transparency}
S.~Forbrigger and Y.~Pan, ``Improving haptic transparency for uncertain virtual
  environments using adaptive control and gain-scheduled prediction,''
  \emph{IEEE transactions on Haptics}, pp. 543--554, 2018.

\bibitem{JRNL:Lawrence_Tele_Passivity}
D.~A. Lawrence, ``Stability and transparency in bilateral teleoperation,''
  \emph{IEEE transactions on Robotics and Automation}, pp. 624--637, 1993.

\bibitem{SimICRA21}
Y.~Sim and J.~Ramos, ``The dynamic effect of mechanical losses of transmissions
  on the equation of motion of legged robots,'' in \emph{2021 IEEE
  International Conference on Robotics and Automation (ICRA)}, 2021, pp.
  2056--2062.

\bibitem{JRNL:L-EXOS}
A.~Frisoli, F.~Salsedo, M.~Bergamasco, B.~Rossi, and M.~C. Carboncini, ``A
  force-feedback exoskeleton for upper-limb rehabilitation in virtual
  reality,'' \emph{Applied Bionics and Biomechanics}, vol.~6, pp. 115--126,
  2009.

\bibitem{CONF:DLRHaptics}
T.~Hulin, K.~Hertkorn, P.~Kermer, S.~Schatzle, J.~Artigas, M.~Sagardia,
  F.~Zacharias, and C.~Preusche, ``The dlr bimanual haptic device with
  optimized workspace,'' in \emph{Proceedings of IEEE International Conference
  on Robots and Automation (ICRA)}, 2011, pp. 3441--3442.

\bibitem{JRNL:HIROIII}
T.~Endo, H.~Kawasaki, T.~Mouri, Y.~Ishigure, H.~Shimomura, M.~Matsumura, and
  K.~Koketsu, ``Five-fingered haptic interface robot: Hiro iii,'' \emph{IEEE
  Transactions on Haptics}, vol.~4, pp. 14--27, 2011.

\bibitem{WEB:Delta}
\BIBentryALTinterwordspacing
{Force Dimension}. (2021) Delta.3. [Online]. Available:
  \url{https://www.forcedimension.com/products/delta}
\BIBentrySTDinterwordspacing

\bibitem{WEB:Phantom}
\BIBentryALTinterwordspacing
{3D Systems}. (2021) Phantom premium 1.5 hf. [Online]. Available:
  \url{https://www.3dsystems.com/haptics-devices/3d-systems-phantom-premium}
\BIBentrySTDinterwordspacing

\bibitem{WEB:HD2}
\BIBentryALTinterwordspacing
{Quanser}. (2021) Hd$^{2}$ high definition hpatic device. [Online]. Available:
  \url{https://www.quanser.com/products/hd2-high-definition-haptic-device/}
\BIBentrySTDinterwordspacing

\bibitem{WEB:Cyberforce}
\BIBentryALTinterwordspacing
{Cyber Glove Systems}. (2021) Cyberforce. [Online]. Available:
  \url{http://www.cyberglovesystems.com/cyberforce}
\BIBentrySTDinterwordspacing

\bibitem{WEB:Inca}
\BIBentryALTinterwordspacing
{Haption}. (2021) Inca. [Online]. Available:
  \url{https://www.haption.com/en/products-en/inca-en.html}
\BIBentrySTDinterwordspacing

\bibitem{CONF:DM2}
M.~Zinn, O.~Khatib, B.~Roth, and J.~K. Salisbury, ``Large workspace haptic
  devices - a new actuation approach,'' in \emph{Proceedings of Symposium on
  Haptic Interfaces for Virtual Environments and Teleoperator Systems}, 2008,
  pp. 185--192.

\bibitem{JRNL:VirtuaPower}
G.~Lee, S.~Hur, and Y.~Oh, ``High-force display capability and wide workspace
  with a novel haptic interface,'' \emph{IEEE/ASME Transactions on
  Mechatronics}, vol.~22, pp. 138--148, 2017.

\bibitem{CONF:Mantis}
G.~Barnaby and A.~Roudaut, ``Mantis: A scalable, lightweight and accessible
  architecture to build multiform force feedback systems,'' in
  \emph{Proceedings of the 32nd Annual ACM Symposium on User Interface Software
  and Technology}, 2019, pp. 937--948.

\bibitem{JRNL:Sunyu_HMI}
S.~Wang and J.~Ramos, ``Dynamic locomotion teleoperation of a reduced model of
  a wheeled humanoid robot using a whole-body human-machine interface,''
  \emph{IEEE Robotics and Automation Letters}, vol.~7, pp. 1872--1879, 2022.

\bibitem{AgileEye}
C.~Gosselin, E.~St.~Pierre, and M.~Gagne, ``On the development of the agile
  eye,'' \emph{IEEE Robotics Automation Magazine}, vol.~3, no.~4, pp. 29--37,
  1996.

\bibitem{BOOK:Modern_Robotics}
K.~M. Lynch and F.~C. Park, \emph{Modern Robotics}.\hskip 1em plus 0.5em minus
  0.4em\relax Cambridge University Press, 2017.

\end{thebibliography}

\end{document}